%% file: main.tex
\documentclass[10pt,twocolumn,letterpaper]{article}

\usepackage[pagenumbers]{cvpr} 

\input{preamble}

\definecolor{cvprblue}{rgb}{0.21,0.49,0.74}
\usepackage[pagebackref,breaklinks,colorlinks,allcolors=cvprblue]{hyperref}

\def\paperID{5826} 
\def\confName{CVPR}
\def\confYear{2025}

\title{Coherent 3D Portrait Video Reconstruction via Triplane Fusion}

\author{Shengze Wang$^\dagger$\\
{\small UNC Chapel Hill}
\and
Xueting Li\\
{\small NVIDIA}
\and
Chao Liu\\
{\small NVIDIA}
\and
Matthew Chan\\
{\small NVIDIA}
\and
Michael Stengel \\
{\small NVIDIA}
\and
Josef Spjut\\
{\small NVIDIA}
\and
Henry Fuchs\\
{\small UNC Chapel Hill}
\and
Shalini De Mello*\\
{\small NVIDIA}
\and
Koki Nagano*\\
{\small NVIDIA}
}

\begin{document}

\twocolumn[{%
\renewcommand\twocolumn[1][]{#1}
\maketitle
\vspace{-3em}
\begin{center}
    *equal contribution
\end{center}
}]

\let\oldthefootnote\thefootnote
\let\thefootnote\relax\footnote{$\dagger$This work was done during an internship at NVIDIA}
\let\thefootnote\oldthefootnote

\input{sec/0_abstract}    
\input{sec/1_intro}
\input{sec/2_related_work}
\input{sec/3_method}
\input{sec/4_results}

\input{sec/5_conclusion}
{
    \small
    \bibliographystyle{ieeenat_fullname}
    \bibliography{main}
}


\end{document}

%% file: preamble.tex
\usepackage[dvipsnames]{xcolor}
\usepackage{multirow}
\usepackage{array} 
\usepackage{graphicx}
\usepackage{keyval}
\usepackage{subcaption}
\usepackage{soul}
\usepackage{tabularx} 
\usepackage{pifont}

\newif\ifdrafting
\draftingfalse 

\ifdrafting
    \newcommand{\ms}[1]{\textcolor{orange}{[Michael: #1]}}
    \newcommand{\tl}[1]{\textcolor{teal}{[Tianye: #1]}}
    \newcommand{\sw}[1]{\textcolor{red}{[Mike: #1]}}
    \newcommand{\yy}[1]{\textcolor{magenta}{[Ye: #1]}}
    \newcommand{\jl}[1]{\textcolor{cyan}{[Jiefeng: #1]}}
    \newcommand{\sd}[1]{\textcolor{blue}{[Shalini: #1]}}
    \newcommand{\kn}[1]{\textcolor{purple}{[Koki: #1]}}
    \newcommand{\todo}[1]{{\color{red}#1}}
    \newcommand{\TODO}[1]{\textbf{\color{red}[TODO: #1]}}
\else
    \newcommand{\ms}[1]{}
    \newcommand{\tl}[1]{}
    \newcommand{\sw}[1]{}
    \newcommand{\yy}[1]{}
    \newcommand{\jl}[1]{}
    \newcommand{\sd}[1]{}
    \newcommand{\kn}[1]{}
    \newcommand{\TODO}[1]{}
    \newcommand{\todo}[1]{}
\fi

\definecolor{bluebox}{RGB}{120, 196, 240} %
\definecolor{graybox}{RGB}{120, 120, 120} %
\definecolor{brownbox}{RGB}{209, 188, 105} %

%% file: sec/0_abstract.tex
\begin{abstract}
Recent breakthroughs in single-image 3D portrait reconstruction have enabled telepresence systems to stream 3D portrait videos from a single camera in real-time, democratizing telepresence. 
However, per-frame 3D reconstruction exhibits temporal inconsistency and forgets the user's appearance. On the other hand, self-reenactment methods can render coherent 3D portraits by driving a 3D avatar built from a single reference image, but fail to faithfully preserve the user's per-frame appearance (\eg, instantaneous facial expression and lighting). As a result, none of these two frameworks is an ideal solution for democratized 3D telepresence. In this work, we address this dilemma and propose a novel solution that maintains both coherent identity and dynamic per-frame appearance to enable the best possible realism. To this end, we propose a new fusion-based method that takes the best of both worlds by fusing a canonical 3D prior from a reference view with dynamic appearance from per-frame input views, producing temporally stable 3D videos with faithful reconstruction of the user's per-frame appearance. Trained only using synthetic data produced by an expression-conditioned 3D GAN, our encoder-based method achieves both state-of-the-art 3D reconstruction and temporal consistency on in-studio and in-the-wild datasets. \url{https://research.nvidia.com/labs/amri/projects/coherent3d}
\end{abstract}

%% file: sec/1_intro.tex
\begin{figure}[ht!]
    \centering
    \includegraphics[width=\linewidth]{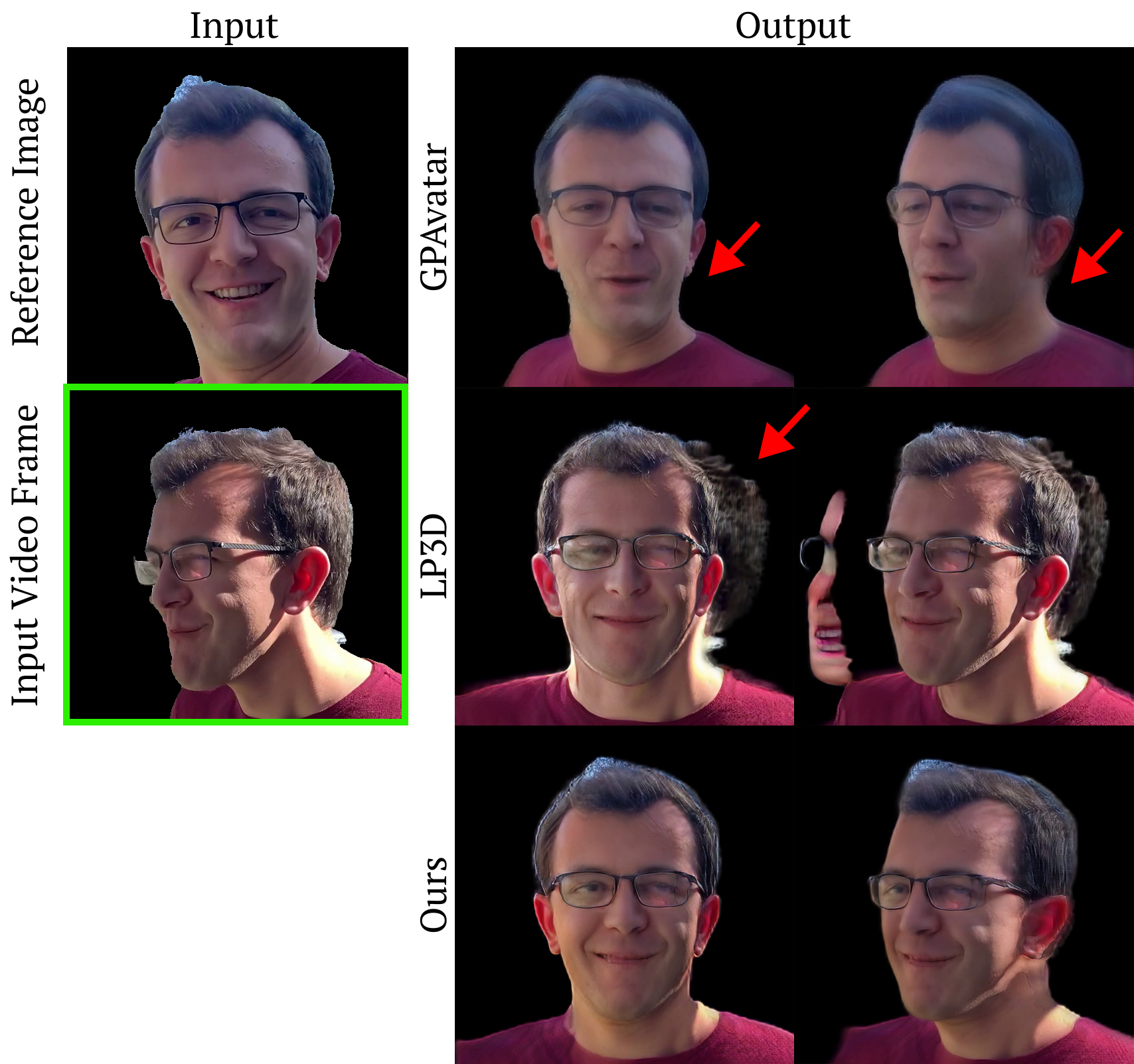}
    \caption{Given a single reference image and a single-view video frame, our method reconstructs the authentic dynamic appearance of the user (e.g., facial expressions and lighting) while producing a temporally coherent 3D video. A previous single-view 3D lifting method (LP3D) that reconstructs the avatar from the video frame on a per-frame basis suffers from distortions and temporal inconsistencies. A portrait reenactment method (GPAvatar) drives the identity in the reference image using the video frame, but fails to capture accurate facial expressions (e.g., smile) and per-frame appearance (e.g., lighting). The output should be compared to the appearance of the per-frame video (green box).}
    \label{fig:teaser}
\end{figure}

\vspace{-5mm}
\section{Introduction}
\label{sec:intro}
\vspace{-2mm}
Telepresence aims at bringing distant people face-to-face and stands out as a particularly compelling application of computer vision and graphics.
Over the last decades, various successful telepresence systems~\cite{raskar1998office, Kauff2002, PixelCodecAvatar, starline, orts-escolano2016holoportation, MAIMONE2012791, jones2009, Kauff2002} have been developed.
However, most employ bulky multi-view 3D scanners or depth sensors to ensure high-quality volumetric per-frame reconstruction. Unlike these classical 3D/4D reconstruction methods, recent AI-based feed-forward 3D lifting techniques~\cite{trevithick2023,bhattarai2024triplanenet} can lift a single RGB image from an off-the-shelf webcam into a neural radiance field (NeRF) representation encoded into a set of triplanes in real-time, paving the way towards making 3D telepresence accessible to anyone~\cite{stengel20233dvc}.

\begin{figure}[ht!]
  \centering
   \includegraphics[width=\linewidth]{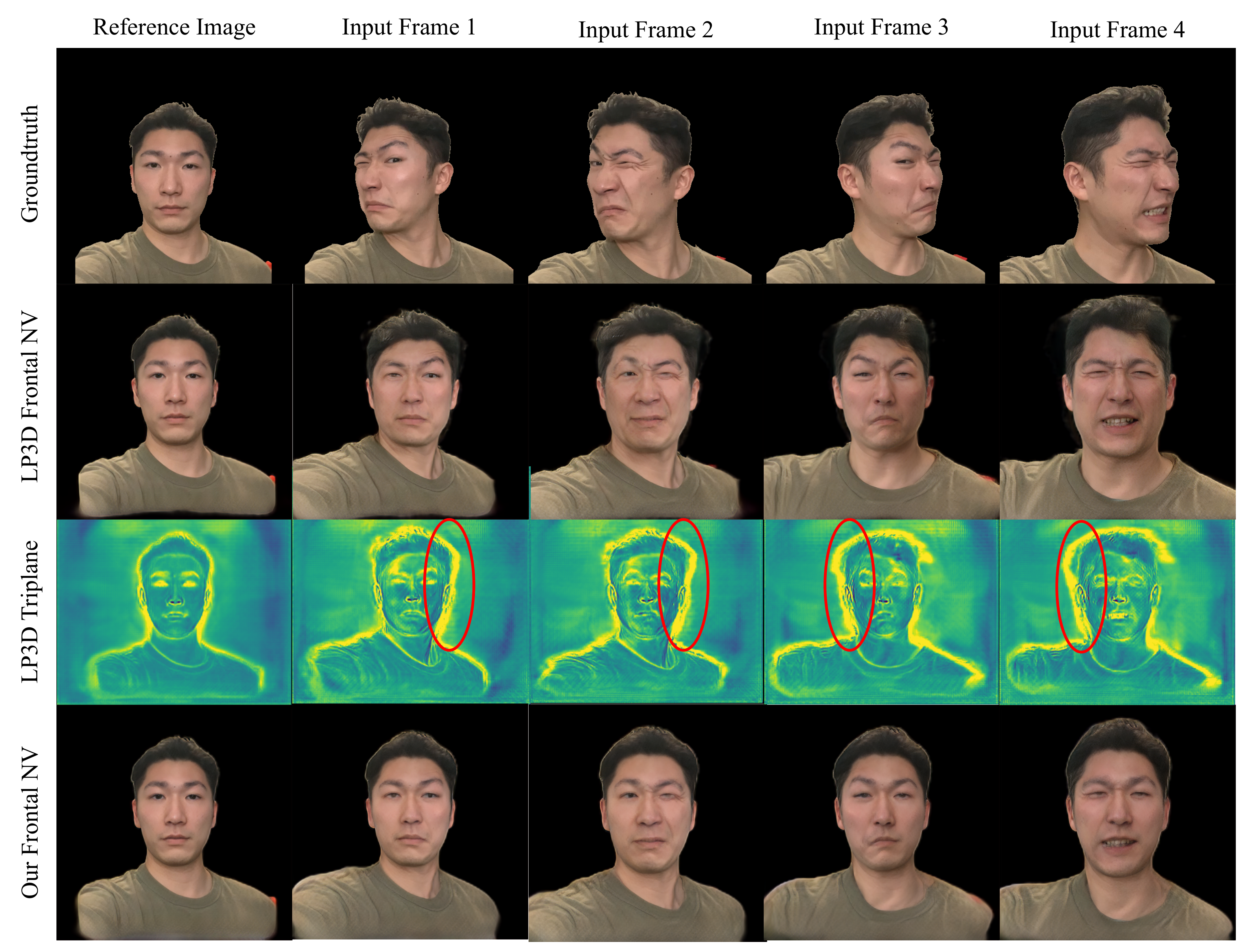}
   \caption{\textbf{View-Dependent Distortion:} \textit{Top}: inputs to our model and LP3D. \textit{Second \& Third Rows:} LP3D's reconstructions varies greatly under challenging viewpoints, showing predictable pattern of artifacts including abnormally strong activations on the side being captured (red circle), as well as geometric distortion along the view direction of the camera. 
   We refer to this phenomenon as "View-Dependent Distortion". \textit{Fourth:} Our method removes such artifacts and achieves better coherence.}
   \label{fig:distortion}
   \vspace{-1em}
\end{figure}

Currently, there are two major paradigms in democratized 3D telepresence solutions from a single-view video: (1) single-view per-frame 3D lifting methods and (2) 3D portrait reenactment, which drives an identity from a reference image using another driving frame, but none of them is an ideal solution. For (1), single-frame-based lifting techniques such as LP3D~\cite{trevithick2023}, have the advantage of faithfully preserving the instantaneous dynamic conditions present in an input video, \eg lighting, expressions, and posture, all of which are crucial to an authentic telepresence experience.
However, single-image reconstruction methods that operate independently on each frame and thus have fundamental limitations for maintaining temporal consistency.
This difficulty stems from the inherent ill-posed nature of single-image-based reconstruction.
In order to render novel views that are significantly far from the input view, the system cannot rely on information present in the input view and hence must \textit{hallucinate} plausible content, which cannot be guaranteed to be consistent across multiple temporal frames
(see the second row in Figs.~\ref{fig:teaser} and \ref{fig:distortion}).
This makes the system susceptible to changes in the lifted 3D portrait's appearance, depending on the user's head pose in the input frame.
In comparison, 3D self-reenactment methods in (2) create an avatar model typically from one or multiple reference frames and use a separate driving video to drive the facial expressions and poses of the avatar.~\cite{tran2023voodoo, chu2024gpavatar, li2023hidenerf,ye2024real3dportrait}. 
While an avatar model allows for temporally consistent results, it often does not faithfully reconstruct the input video's dynamic conditions such as change of lighting, accessories, or hair motion.
Moreover, reenactment methods often struggle to authentically reconstruct the accurate expressions of the user because the expression control is not precise enough (see the first row in Fig.~\ref{fig:teaser}).

In this work, we address this dilemma of democratized 3D telepresence approaches, and propose a novel solution to the problem of simultaneously maintaining temporal stability while preserving real-time dynamics of input videos for 3D lifting-based human telepresence applications.
Our proposed solution is a fusion-based approach that leverages the stability and accuracy of a canonical 3D prior, and also captures the diverse deviations from the prior from per-frame observations (see the third row in Fig.~\ref{fig:teaser}).

Our model first uses LP3D~\cite{trevithick2023} to construct a canonical\footnote{We find that the 3D lifting from a near frontal reference view is reliable, hence use this as a canonical 3D prior. See Fig.~\ref{fig:distortion} first column.} triplane prior from a (near) frontal image of the user, which can be casually captured or extracted from a video.  
During video reconstruction, our model lift each input frame into a raw triplane, which is then fused with the canonical triplane (see Fig.~\ref{fig:overview}).
When the head pose of the input image is oblique, artifacts and identity distortions may be present in its lifted triplane (see "LP3D" in Fig.~\ref{fig:teaser} and \ref{fig:distortion}).
Hence we propose an undistorter module, which learns to undistort the raw instantaneous triplane to more closely match the structure of the correctly-structured canonical triplane.
We then propose a fuser module, which learns to densely align the undistorted raw triplane to the reference triplane and then fuse the two in a manner that incorporates personalized details such as tattoos or birthmarks present in the reference triplane, while preserving dynamic lighting, expression and posture information from the input raw triplane.

We summarize our contributions as follows:
\begin{itemize}
    \item We contribute a novel triplane fusion method that combines the dynamic information from per-frame triplanes with a canonical triplane extracted from a reference image. 
    Trained only using a synthetic multi-view video dataset, our feed-forward approach generates 3D portrait videos that demonstrate both temporal consistency and faithful reconstruction of the dynamic appearance  of the user (\eg lighting and expression), whereas prior solutions can only achieve one of the two properties.     
    \item We propose a new framework to evaluate single-view 3D portrait reconstruction methods using multi-view data and gain insight to the method's reconstruction quality and robustness.
    \item We present evaluations on both in-studio and in-the-wild datasets and demonstrate that our method achieves state-of-the-art performance in terms of temporal consistency and reconstruction accuracy. 
\end{itemize}

%% file: sec/2_related_work.tex
\vspace{-3mm}
\section{Related Work}

\noindent\textbf{2D portrait reenactment.}
Given a single or a few reference portrait images and a driving video, recent talking-head generators can reenact 2D portraits by transferring the facial expressions and poses from the driving video onto facial portraits~\cite{wang2021facevid2vid, Siarohin_2019_NeurIPS,Zakharov_2019_ICCV,Zakharov20,zhang2022metaportrait,doukas2020headgan,Drobyshev22MP,hong2022depth,tps2022,wang2022latent,StyleHeat2022}. However, being 2D, they cannot be rendered from novel viewpoints, which is crucial for 3D telepresence.

\noindent\textbf{3D-aware portrait generation and reenactment.}
Some recent works use deformable volumetric implicit radiance fields~\cite{mildenhall2020nerf,park2021nerfies,pumarola2020d} or Gaussian splatting~\cite{kerbl3Dgaussians} combined by 
3DMMs to reconstruct a photorealistic and animatable volumetric head avatar~\cite{Gafni_2021_CVPR,cao2022authentic,zheng2022imavatar,athar2022rignerf,xu2023gaussianheadavatar,qian2023gaussianavatars,saito2024rgca}. However, they require extensive data captures from videos or multiview cameras and person-specific training. Others use large-scale video datasets and learn a disentangled triplane 3D~\cite{eg3d2022} for 3D facial reenactment in a feedforward fashion~\cite{ye2024real3dportrait, chu2024gpavatar,tran2023voodoo, Li2023Oneshot,ma2023otavatar, li2023hidenerf,NOFA2023Yu}. They construct a canonical 3D head from a reference image (often a neutral frame), and use facial expressions and head poses extracted from a separate driving video to animate it. As such, fine-grained facial expressions may not be captured due to errors in disentanglement. Most importantly, these reenactment methods fail to preserve the dynamic appearance of users (e.g., person-specific wrinkles or lighting) across time.

\begin{figure*}[ht!]
  \centering
    \includegraphics[width=0.9\textwidth]{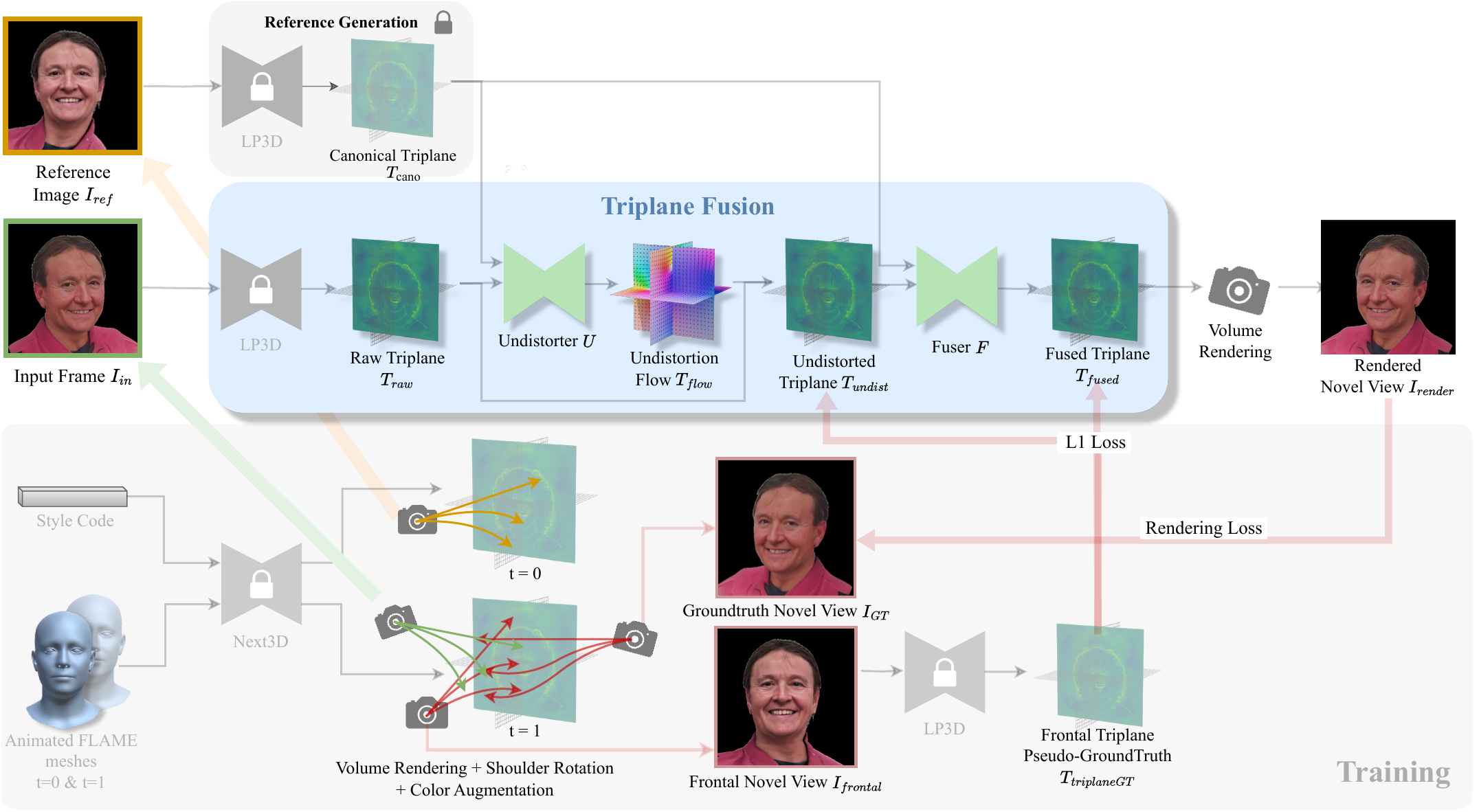}
   \caption{\textbf{Overview}. Given a (near) frontal reference image and an input frame, we reconstruct a canonical triplane and a raw triplane, respectively, using LP3D~\cite{trevithick2023} (Sec.~\ref{sec:lp3d}). Next, we combine these two triplanes through a Triplane Fusion module (blue box) that ensures temporal consistency while preserving realtime dynamics (\textit{e.g.}, lighting and shoulder pose) (Sec.~\ref{sec:undistorter} and Sec.~\ref{sec:fuser}). Our model is trained with only synthetic video data generated by a 3D GAN~\cite{sun2023next3d}, with carefully designed augmentations to preserve shoulder motion and lighting (Sec.~\ref{sec:data}).
   }
   \label{fig:overview}
   \vspace{-1em}
\end{figure*}

\noindent\textbf{3D GAN inversion.} 
By combining GANs~\cite{goodfellow2014generative} and neural volume rendering~\cite{mildenhall2020nerf}, recent breakthroughs in 3D-aware GANs~\cite{eg3d2022,orel2022stylesdf,gu2021stylenerf,deng2022gram,xiang2022gramhd,epigraf,Zhou2021CIPS3D,zhang2022mvcgan,xu2021volumegan,rebain2022lolnerf,xu2022pv3d} demonstrate unsupervised learning of photorealistic 3D heads from in-the-wild 2D images. Notably, EG3D proposes a triplane representation~\cite{eg3d2022}, which is efficient and compact. Next3D~\cite{sun2023next3d} extends EG3D to create 3D portrait videos controlled by 3DMM facial expression and pose parameters, which we also use to create our synthetic multiview video training data. 
Once these 3D head priors are trained, they can be used to perform single-view 3D reconstruction using GAN inversion to lift a portrait to 3D~\cite{ko20233d3dganinversion,sun2022ide,lin20223dganinversion,Fruehstueck2023VIVE3D}, manipulate the 3D avatar~\cite{hong2021headnerf,sun2022ide,zhuang2022mofanerf}, or 3D personalization~\cite{buehler2023preface, qi2023my3dgen}. Since GAN inversion is time consuming, recent works ~\cite{trevithick2023,bhattarai2024triplanenet} propose an encoder-based solution to lift a single facial image into a triplane. However, for a video, they lift every frame independently, and exhibit temporal inconsistency -- a key limitation to creating a practical 3D telepresence system. To enhance single-frame-based based 2D-to-3D encoders for human heads, \textit{e.g.}, LP3D~\cite{trevithick2023}, we propose a triplane-fusion-based method, which improves their temporal stability while preserving temporal dynamics across time.

%% file: sec/3_method.tex
\section{Definitions}
For telepresence, we aim to create more temporally coherent 3D portrait videos from an input 2D video without test-time optimization, while preserving its unique temporal dynamics (lighting, expressions, shoulder pose, \textit{etc.}). 
We define the terminology for our task. 
We call a current facial video frame an ``\textbf{input frame}" into our system and convert it to a 3D portrait. 
Additionally, we assume that a near-frontal ``\textbf{reference image}" of the subject is encoded into a ``\textbf{canonical triplane}" and used to stabilize the 3D video generation. Lastly, we refer to the viewpoint of the input frame relative to the user's head as the ``\textbf{input viewpoint}".

\section{Method}
We aim to reconstruct coherent 3D portrait videos from a monocular RGB video. An overview of our method is illustrated in Fig. \ref{fig:overview}.
To improve temporal consistency and reconstruct occluded parts of the face, we use an additional near-frontal reference image obtained from the same video or a selfie capture and lift it into a canonical triplane. 
We convert each incoming video input frame into a raw triplane 
using a pre-trained LP3D encoder (Sec.~\ref{sec:lp3d}). 
Then, the Triplane Undistorter (Sec. \ref{sec:undistorter}) removes view-dependent distortions and artifacts from it using the canonical triplane and produces an undistorted triplane. 
Finally, to recover regions that are occluded in the input frame, our Triplane Fuser (Sec. \ref{sec:fuser}) combines it with the canonical triplane to generate the final coherent triplane.

\subsection{Background: 3D Portrait from a Single Image}
\label{sec:lp3d}
\indent LP3D\cite{trevithick2023} performs photorealistic 3D portrait reconstruction with a feedforward encoder to convert an RGB image into a triplane $\textbf{T} \in \mathbb{R}^{3\times 32\times256\times256}$, which can be volume rendered to an RGB image from any viewpoint.
LP3D can run in real-time and has been developed into a complete realtime telepresence system~\cite{stengel20233dvc}. We use LP3D to lift a 2D face into triplanes with a slightly modified version trained on larger face crops containing shoulders and a camera estimator to recover the input image's camera parameters $M \in \mathbb{R}^{25}$. It performs better than the original version (Table~\ref{tab:rebuttal_tab}).

\subsection{Generating Synthetic Dynamic Multiview Data}
\label{sec:data}
We use Next3D~\cite{sun2023next3d} to generate animated 3D portraits and rendered images from them as groundtruth data to train our model. We label the 
FFHQ~\cite{karras2019style} dataset with 2D landmarks and FLAME~\cite{FLAME:SiggraphAsia2017} expression coefficients using 
DECA~\cite{feng2021learning}. During training, we sample a pair of random FLAME coefficients from FFHQ and provide to Next3D, with a single random $z$ identity code to generate a pair of triplanes for $t=0$ and $t=1$ with different expressions of the same person (Fig.~\ref{fig:overview}).  
We render the triplane at $t=0$ into a near-frontal reference image (Fig.~\ref{fig:overview} bottom) and the triplane at $t=1$ into 3 images: the input frame, the groundtruth at another sampled novel viewpoint, and a frontal novel view image.
The later two are only used for supervision.
To learn to fuse images under different lighting conditions, we also apply two separate color space augmentations to the rendered images containing alterations to brightness, contrast, saturation, and hue. 

\noindent\textbf{Shoulder augmentation.}
It is important that 3D portraits captures dynamic shoulder movement to convey body language and achieve eye contact in telepresence. 
Next3D does not provide control over shoulder posture. So, we warp camera rays during volume rendering to simulate shoulder movement in the rendered image without having to modify the Next3D triplane. In Table.~\ref{tab:rebuttal_tab} we show quantitative improvements from using shoulder augmentation. Please see the supplement for more details. 
 
\noindent\textbf{Pseudo-groundtruth triplanes.} \label{sec: pseudo_gt}
As a result of the shoulder augmentation, the Next3D triplanes and their 2D renderings are now different.
Thus, the Next3D triplanes cannot be used as direct supervisory signals. 
To mitigate this, we leverage the fact that LP3D generates reasonably accurate triplanes from frontal view images.
We use a frozen LP3D to predict pseudo-groundtruth triplanes $T_{frontalGT}$ from the frontal novel view for $t=1$ (Fig.~\ref{fig:overview} bottom). 

\subsection{Removing Distortion and Preserving Identity}
\label{sec:undistorter}
LP3D's reconstruction quality is highly dependent on the viewpoint of the input frame. 
When a person is captured from the side, LP3D often produces incorrect identity and stretching distortion (\textit{e.g.}, in the ``Input Frame 1" and ``Input Frame 2" columns in Fig.~\ref{fig:distortion}). 
This can be ascribed to the inherent ambiguity of single-image reconstruction.
However, LP3D works well with frontal views with more complete identity information and less occlusion than with side views (\textit{e.g.} the ``Reference Image" column in Fig.~\ref{fig:distortion}).
Therefore, for coherent temporal reconstruction, we aim to reduce the single-image ambiguity by leveraging an extra near-frontal reference image, which is encoded as a canonical triplane $T_{cano}$. 
We use a single image as the reference to keep the user interface simple and find it to be sufficient to improve temporal coherence.

To correct the distortion in an input raw triplane $T_{raw}$ using the canonical triplane $T_{cano}$ as reference, we devise a Triplane Undistorter $U$ (Fig.~\ref{fig:overview}):
\begin{equation}
    \vspace{-1mm}
    U(T_{raw}, T_{cano})=T_{undist} \in \mathbb{R}^{3\times32\times256\times256}\ .
    \vspace{-1mm}
\end{equation}
Since the distortion in each triplane is a 2D warping artifact, one can reverse this warping by predicting 2D undistortion warping for the three planes.
To achieve this, our Undistorter $U$ adopts the SPyNet\cite{ranjan2017optical} architecture to predict 2D correction warping $T_{corr} \in \mathbb{R}^{3\times2\times256\times256}$ for the three planes.
Then, the Undistorter corrects the raw triplane $T_{raw}$ by warping it based on the estimated 2D movement $(\Delta u, \Delta v)$ at each feature pixel in $T_{corr}$. 
\begin{gather}
    T_{corr} = SPyNet(T_{raw}, T_{cano}),\\
    T_{undist}=Warp(T_{raw}, T_{corr}).
    \vspace{-1mm}
\end{gather}
It is important to note that, although SPyNet is originally designed to warp a source image to a target image, SPyNet is trained to function differently during undistortion correction.
We later show in Table.~\ref{tab:rebuttal_tab} that optical flow alignment leads to significant artifacts and destroys the reconstruction. 
This is because an ideal optical flow network would warps $T_{raw}$ to aligns it to $T_{cano}$, making the two identical.
However, in our case, the goal is to correct $T_{raw}$ such that it has the same identity as the canonical triplane $T_{cano}$, but \emph{preserves the different dynamic information such as expressions and lighting}.
Therefore, instead of warping towards the canonical triplane $T_{cano}$, the input triplane $T_{raw}$ should be warped towards the groundtruth triplane $T_{triplaneGT}$ (Fig.\ref{fig:overview} lower-right), which is not available during test time.
Therefore, our undistorter $U$ merely uses $T_{cano}$ as the identity conditioning to predict a correction warping to $T_{raw}$ but not the target for warping. 
The correction warping is supervised by the consistency between $T_{undist}$ and pseudo-groundtruth triplane $T_{triplaneGT}$ (Sec.~\ref{sec:data}) via a triplane loss:
\begin{equation}
    \vspace{-1mm}
    L_{undist} = L_1(T_{undist}, T_{triplaneGT}).
\end{equation}

\subsection{Reconstructing Occluded Regions through Triplane Fusion}
\label{sec:fuser}
As the user moves around in the video, different parts of their head become occluded. 
To recover occluded areas in the input frame and further stabilize the subject's identity across the video, our Fuser $F$ enhances the reconstruction by utilizing the canonical triplane $T_{cano}$, where the currently occluded areas are often visible.
Therefore, it is important for the Fuser $F$ to identify and recover the occluded regions while preserving information from visible regions in $T_{undist}$.
To accomplish this, we thus use a 5-layer ConvNet-based visibility estimator $V$ to predict a visibility triplane for the input frame by estimating a visibility triplane $V(T_{raw})=T_{raw}^{vis} \in \mathbb{R}^{3\times 1\times128\times128}$, \ie one visibility map for each plane.
$T^{vis}_{raw}$ is undistorted alongside $T_{undist}$ to produce $T_{undist}^{vis}$. 
We also predict a visibility triplane for the canonical triplane as $T^{vis}_{cano}=V(T_{cano})$.
Finally, the Fuser $F$ produces the fused triplane $T_{fused}$ by combining information from the undistorted input tranplane $T_{undist}$, its visibility triplane $T_{undist}^{vis}$, the canonical triplane $T_{cano}$, and its visibility triplane $T_{cano}^{vis}$. 
\begin{equation}
    T_{fused} = F(T_{undist}, T_{undist}^{vis}, T_{cano}, T_{vis}^{cano})
\end{equation}
In this way, Fuser $F$ preserves visible facial regions in $T_{raw}$ and can recover the occluded regions using the canonical triplane $T_{cano}$.

To train the visibility predictor $V$, we calculate the visibility loss $L_{vis}$ as the $L_1$ distance between the predicted visibility triplanes ($T_{raw}^{vis}$ and $T_{cano}^{vis}$) versus the groundtruth visibility triplanes ($T_{raw}^{visGT}$ and $T_{cano}^{visGT}$):
\begin{equation}
L_{vis} = L_1(T^{raw}_{vis}, T^{raw}_{visGT}) + L_1(T^{cano}_{vis}, T^{cano}_{visGT}).
\end{equation}
The visibility triplane $T_{visGT}$ contain 1 for pixels that are visible, and 0 otherwise. 
The generation of the groundtruth visibility triplanes $T^{raw}_{visGT}$ and $T^{cano}_{visGT}$ is discussed in the supplementary.

To supervised the Fuser $F$, we calculate the fusion loss $L_{fusion}$ as the $L_1$ loss between the fused triplane $T_{fused}$ and the pseudo-groundtruth triplane $T_{frontalGT}$. To highlight the currently occluded region during training, we also upweight the occluded region using an occlusion mask $T_{occMask} \in \mathbb{R}^{3\times 1\times 256\times 256}$ (please refer to the supplementary for the calculation of occlusion mask):
\begin{align}
T_{diff} = |T_{fused} - T_{triplaneGT}|\\
\scalebox{0.8}{$
L_{fusion} = Mean(T_{diff}) + \dfrac{T_{diff}\cdot T_{visGT}}{|T_{visGT}|}  + \dfrac{T_{diff}\cdot T_{occMask}}{|T_{occMask}|} 
$}
\end{align}
We use the Recurrent Video Restoration Transformer (RVRT)~\cite{rvrt} as the backbone of our Fuser $F$ because of its memory efficiency. 
We find that the final summation skip connection in RVRT prevents effective learning.
This is because the original RVRT was designed to correct local blurriness and noises in a corrupted RGB video, whereas our triplane videos exhibit structural distortion on a much larger scale and the summation skip connection thus limits the model's ability to correct the general structure.
We thus replace the summation with a small 5-layer ConvNet.

Lastly, note that both the Undistorter $U$ and the Fuser $F$ consist of 3 separate copies, one for each of the 3 planes because we find that processing all three planes jointly leads to collapse to 2D (please see supplementary for the visualization and analysis).

\subsection{Training Losses}
Our loss function is the summation of four loss terms that provide two types of supervision: (a) direct triplane space guidance used to supervise the undistortion process in the Undistorter $U$, the visibility prediction process, and the fusion process in the Fuser $F$; and (b) image space guidance for overall learning of high-quality image synthesis:
\vspace{-5pt}
\begin{align}
L = w_{undist} L_{undist} + w_{fusion} L_{fusion} \notag \\
     + w_{vis} L_{vis} + w_{render} L_{render} .
\end{align}
$w_{undist}$, $w_{viz}$ , $w_{fusion}$, and $w_{render}$ are scalar weights for the different loss terms. \noindent $L_{render}$ is calculated as the perceptual loss $L_{LPIPS}$ between the groundtruth novel view $I_{GT}$ and the rendered novel view $I_{render}$:
\begin{equation}
L_{render} = L_{LPIPS}(I_{GT}, I_{render}) 
\end{equation}

%% file: sec/4_results.tex
\section{Results}
As discussed before, current methods like LP3D~\cite{trevithick2023} can overfit to the input viewpoints, but exhibits significant artifacts when synthesizing novel viewpoints for challenging input views like a profile picture.
Therefore, we need to evaluate the methods by examining their reconstruction across multiple viewpoints instead of only from the input view as done previously.

\subsection{Metrics} \label{sec:metrics}
We measure the identities accuracy as the ArcFace\cite{deng2018arcface} cosine distance between the $I_{render}$ and $I_{GT}$. 
To measure the accuracy of reconstructed expressions, we use the NVIDIA Maxine AR SDK to measure the $L_2$ distance between expression coefficients $e_{render}$ of the rendered image and $e_{GT}$ of the groundtruth.

\vspace{1em}
\noindent\textbf{Multi-view evaluation of single-view reconstruction.} 
Due to the lack of 3D ground-truth for real-world data, prior methods are often evaluated on the input view reconstruction task using quantitative metrics like PSNR, whereas the novel view synthesis task often relies on visual assessments.
However, evaluating a reconstruction using only a single viewpoint can lead to ambiguities and inaccurate conclusions. 
For example, if the evaluation is only performed using the input viewpoint, then a method can overfit to the input view to achieve high numeric scores even if its reconstruction is highly inaccurate when rendered from novel viewpoints.
Moreover, single-view reconstruction methods can be heavily affected by the choice of input viewpoints. As shown in Fig.~\ref{fig:distortion}, different input views can lead to very different reconstructions.
Therefore, there are two variables crucial to the evaluation of single-view reconstruction methods: 
the input viewpoint and the evaluation viewpoint.
We thus propose new multi-view metrics that evaluate a model across different input-evaluation viewpoint combinations.
Using these new metrics, a method can only achieve high numeric performances when it consistently generates high-quality reconstructions regardless of the choice of input or evaluation viewpoints:

\vspace{1em}
\noindent\textbf{Overall synthesis quality.} 
Given $N$ views in the dataset, we evaluate a method's average performance across different input-evaluation viewpoint combinations. More specifically, at each frame, each of the $N$ cameras is used as the input viewpoint to produce $N$ reconstructions in total, and each of the $N$ reconstruction is rendered and evaluated on the $N$ viewpoints, resulting in an $N\times N$ score matrix (please refer to the supplementary for visualization of the score matrix). 
We use $N=8$ views in the NeRSemble\cite{Nersemble} dataset.
Thus, for a test sequence with $T$ frames, we generate a spatial-temporal score matrix $\textbf{S}^{T\times 8\times 8}$ for each of the metric (see the supplement for example visualization):
\vspace{-2mm}
\begin{align}
\scalebox{0.8}{$\textbf{S}_{t,i,j} = Metric(\textbf{I}_{render}^{t,i,j}, \textbf{I}_{GT}^{t,j}),1\leq i, j\leq N , 1\leq t\leq T$},\\
    s = Mean(\{\textbf{S}_{t,i,j}\}).
    \vspace{-1mm}
\end{align}
where $Metric(\cdot)$ can be LPIPS, PSNR, $ID$, and $Expr$. 
$\textbf{I}_{render}^{t,i,j}$ is the image rendered using camera $i$ as the input frame and camera $j$ as the output rendering view at frame $t$. $\textbf{I}_{GT}^{t,j}$ is the groundtruth frame captured by camera $j$ at frame $t$.
The Overall Synthesis Quality $s$ is thus the average over all score entries in $\textbf{S}$.
For a dataset of multiple test sequences, the final Overall Synthesis Quality is the average score of all sequences.

\vspace{1em}
\noindent\textbf{Novel view synthesis (NVS) quality.} 
Novel View Synthesis Quality $s_{NV}$ is the average over all scores corresponding to novel view synthesis, \ie, the input view $i$ is different from output rendering view $j$:
\begin{equation}
s_{NV} = Mean(\{\textbf{S}_{t,i,j} | i\neq j, 1\leq t\leq T \}) .
\end{equation}

Additionally, it is also important to measure whether a method can authentically reconstruct dynamic real-life conditions in the video such as changes in lighting and shoulder poses. However, there is no existing multi-view in-the-wild portrait video dataset to support the evaluation of view synthesis quality. We thus qualitatively evaluate the methods on challenging in-the-wild portrait videos. Please see supplementary materials for image examples and video results.

\subsection{Comparisons}
\begin{figure*}[ht]
  \centering
    \includegraphics[width=\textwidth]{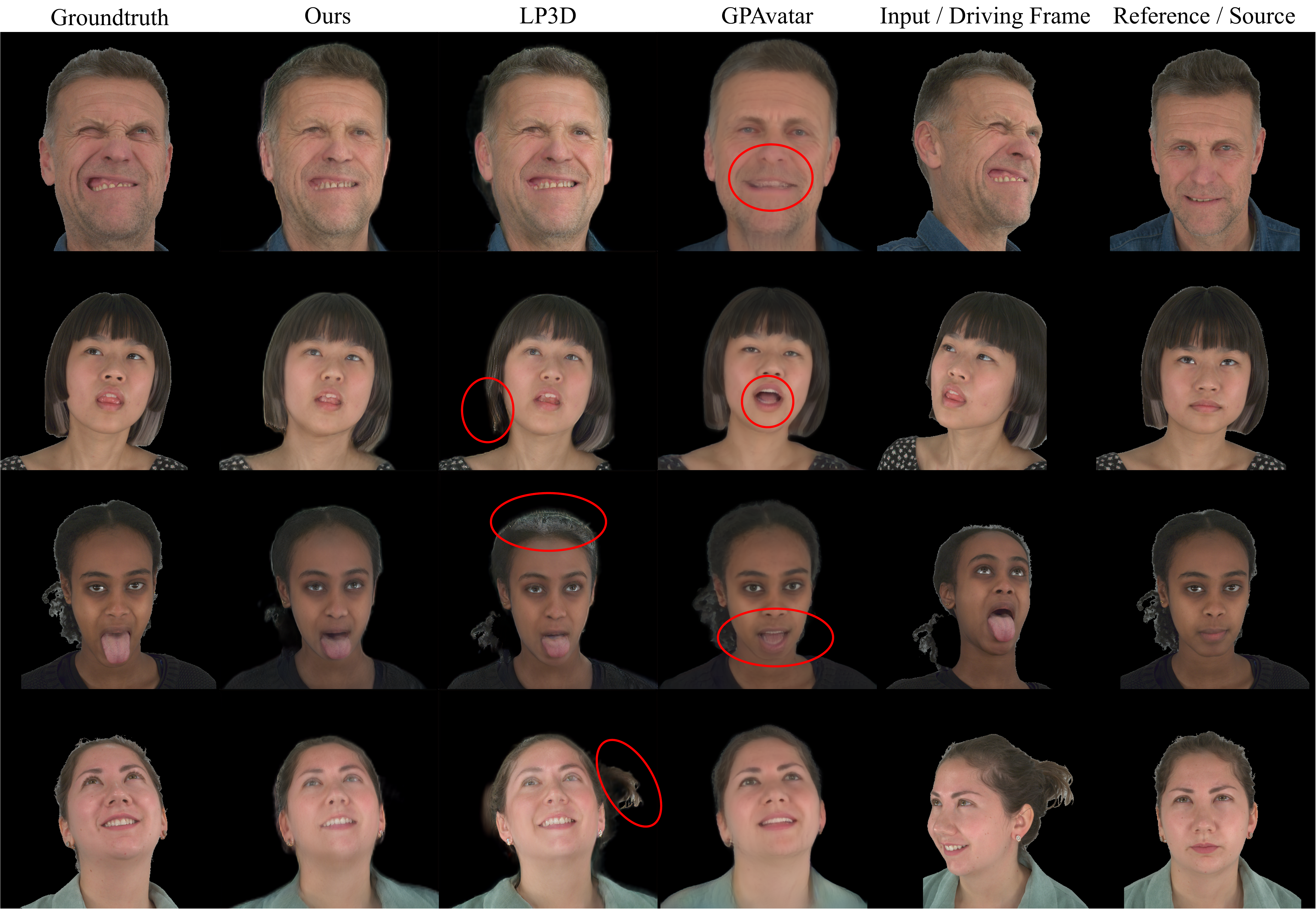}
   \caption{\textbf{Visual comparisons with baseline methods.} Our method strikes a balance between coherent reconstruction and faithful dynamic conditions like expressions. LP3D (third column) exhibits inconsistencies in identities, hairstyles, and artifacts (red circles). GPAvatar (fourth column) fails to capture challenging expressions (first row), new information not present in the reference image, (the stuck-out tongue in second and third rows), and identity of the person (last row).
   }
   \label{fig:comparison}
\end{figure*}

\noindent\textbf{Baselines.} We compare our method with recent methods from 3 categories:

\textit{Reconstruction:} We evaluate LP3D\cite{trevithick2023} using the above protocol. We provide LP3D with the image from the input viewpoint and evaluate on all 8 viewpoints from the NeRSemble dataset. 

\textit{Reenactment:} Li \etal\cite{Li2023Oneshot} 
is able to reconstruct 3D portraits into a triplane from a reference image without test-time optimization, and they drive the reeconstruction via the frontal rendering of a 3DMM that modifies the expression in the original triplane. Concurrent to our work, GPAvatar\cite{chu2024gpavatar}  reconstructs 3D portraits by leveraging multiple source images and driving them through a FLAME\cite{FLAME:SiggraphAsia2017} mesh model. We test both methods in the self-reenactment setting. We use the first frame of the frontal camera in each NeRSemble test sequence as the reference image, and we drive it using videos of all 8 viewpoints. We evaluate GPAvatar using the same evaluation protocol as our method and LP3D. We evaluated Li \etal~\cite{Li2023Oneshot}'s approach using the input views as the only evaluating views, which are computed by the original authors, instead of all 8 views.

\textit{Inversion:} We also evaluate VIVE3D\cite{Fruehstueck2023VIVE3D}, which is a state-of-the-art 3D GAN inversion method for videos, and it can also perform semantic video editing. To perform inversion and evaluation on NeRSemble, VIVE3D's 3D GAN is first personalized using 3 frames from the input viewpoint video before inverting and rendering the reconstructed video from all 8 viewpoints.

Unfortunately, each of the above methods use different croppings of the face. We standardize the evaluation by re-cropping all methods to our cropping protocol, which is the largest of all. Please see the supplement for an additional table, where we evaluate the methods using different croppings around the face and arrive at conclusions consistent with Table.~\ref{table:split_column}.

\noindent\textbf{Dataset.}
We quantitatively evaluate the methods on the NeRSemble~\cite{Nersemble} dataset, which is a high-quality multi-view portrait video dataset recorded with 16 calibrated time-synchronized cameras in a controlled studio environment. 
The images are captured at 7.1 MP resolution and 73 frames per second. 
There are 10 recordings in the test set, capturing a total of 10 individuals performing different expressions. 
NeRSemble provides us with the ability to evaluate the Overall Synthesis Quality and NVS Quality using the different input-evaluation viewpoint combinations. 
One of the 10 test sequences involves severe facial occlusion from hair that causes most of the methods' face trackers to fail for significant portions of the recording for many of the viewpoints. 
We thus leave out that sequence because the results would not be a reliable assessment of quality. 
We also use 8 roughly evenly separated cameras out of all 16 cameras during the evaluation.

\noindent\textbf{Quantitative results.} As mentioned before, we evaluate the methods using different input-evaluation viewpoint combinations, providing robust multi-view estimation for each of the metrics. Table.~\ref{table:split_column} shows that our model achieves state-of-the-art performance across all metrics versus recent works. 
Notably, LP3D is heavily affected by the input viewpoint, and our method is able to better preserve subject identity and expression (see Fig.~\ref{fig:distortion} and \ref{fig:comparison}). 
On the other hand, the reenactment methods struggle to capture authentic expressions because of the use of morphable face models, which have limited expressiveness. Moreover, they cannot faithfully reconstruct dynamic conditions (e.g. the stuck-out tongue in the second and third rows of Fig.~\ref{fig:comparison}) because they solely rely on information present in the source/reference images and do not incorporate new per-frame information. 
On the other hand, our method faithfully captures dynamic conditions and coherent reconstruction at the same time.

\begin{table}[ht]
\centering
\fontsize{8}{13}\selectfont 
\setlength{\tabcolsep}{2pt} 
\begin{tabularx}{\linewidth}{lccccccc}
\toprule
 Method & Type & Expr$\downarrow$ & ID$\downarrow$ & \multicolumn{2}{c}{Synthesis Quality} & \multicolumn{2}{c}{NVS Quality} \\
& & & & PSNR$\uparrow$ & LPIPS$\downarrow$ & PSNR$\uparrow$ & LPIPS$\downarrow$\\
\hline
Li \etal \cite{Li2023Oneshot} & reenact & 0.266 & 0.241 & 18.573 & 0.255 & 18.202 & 0.262 \\  
GPAvatar\cite{chu2024gpavatar} & reenact & 0.204 & \underline{0.207}  & 21.949 & 0.233 & \underline{21.949} & \underline{0.2334} \\
VIVE3D\cite{Fruehstueck2023VIVE3D} & invert & 0.290 & 0.395 & 18.577 & 0.259 & 18.145 & 0.271 \\  
LP3D\cite{trevithick2023} & recon & \underline{0.168} & 0.215 & \underline{22.331} & \underline{0.223} & 21.525  & 0.237\\
Ours & recon & \textbf{0.158} & \textbf{0.187}  & \textbf{22.770} & \textbf{0.219} & \textbf{22.440} & \textbf{0.224}\\ \bottomrule
\end{tabularx}
\caption{\textbf{Comparison on Nersemble~\cite{Nersemble}:} Our evaluation protocol (Sec.~\ref{sec:metrics}) utilizes multi-view groundtuth to evaluate each model. Under this robust evaluation, our method achieves state-of-the-art performance across all metrics. Our method achieves the best view synthesis accuracy and robustness to input viewpoints ("Synthesis Quality" $\&$ "NVS Quality") while accurately capturing the identity and expression.}
\label{table:split_column}
\end{table}

\begin{table}[t]
\centering
\fontsize{8}{13}\selectfont 
\setlength{\tabcolsep}{2.2pt} 
\begin{tabularx}{\linewidth}{p{1.2cm} c c c c c c c c}
\toprule
Method & $U$ & $F$ & PSNR$\uparrow$ & LPIPS$\downarrow$ & \multicolumn{2}{c}{IVV} & \multicolumn{2}{c}{NVV}\\
& & & & & PSNR$\downarrow$ & LPIPS$\downarrow$ & PSNR$\downarrow$ & LPIPS$\downarrow$ \\
\hline
LP3D\cite{trevithick2023} & 	\ding{55} & \ding{55} & \underline{22.331} & 0.223 & 1.025 & 0.015 & 2.200 & 0.053 \\
\hline
Ours & \checkmark & \ding{55} & 22.196 & 0.221 & 0.907 & 0.009 & 1.699 & 0.038 \\
 & \ding{55}& \checkmark & 22.265 & \underline{0.223} & \underline{0.559} & \underline{0.006} & \textbf{1.315} & \textbf{0.029} \\
 & \checkmark & \checkmark & \textbf{22.769} & \textbf{0.219} & \textbf{0.245} & \textbf{0.005} & \underline{1.383} & \underline{0.037} \\ \bottomrule
\end{tabularx}
\caption{\textbf{Undistorter and Fuser Ablations.} We test two variations of our models (1) Undistorter-only (row 2), and (2) Fuser-only (row 3). We show that simply adding each component does not lead to improvement. However, they complement each other and substantially improves the accuracy to the reconstruction (NVV - Novel View Variation) as well as the robustness to challenging input viewpoints (IVV - Input View Variation).}
\label{table:ablation}
\vspace{-0.7em}
\end{table}

\noindent\textbf{Ablations.} We compare 5 model variations: (1) the original LP3D (Table.~\ref{tab:rebuttal_tab}, row 1), (2) Ours with optical flow instead of undistorter (Table.~\ref{tab:rebuttal_tab}, row 3), (3) Ours without shoulder augmentation (Table.~\ref{tab:rebuttal_tab}, row 4), (4) Ours with only the Triplane Undistorter (Table.~\ref{table:ablation}, row 2), and (5) Ours with only the Triplane Fuser $F$ (row 3)

To measure robustness to input viewpoints and the consistency of novel view rendering across evaluation viewpoints, we develop two new metrics: 

\noindent(a) \textbf{Novel View Variation (NVV)} 
We evaluate how much a method's reconstruction quality varies across different evaluation views. 
We quantify this as the standard deviation of performance across the $N$ evaluating views using the same input view, \ie horizontal rows $1\leq i\leq N$ of the score matrix $S$ (Tab.~\ref{table:ablation} second column from the right): 
\begin{align}
NVV = Mean(\{Stddev(\{\textbf{S}_{t,i,j} | i\neq j\})\} \notag \\
    | 1\leq i\leq N, 1\leq t\leq T\}) .
\end{align}
(b) \textbf{Input View Variation (IVV)} 
We measure how much a method's reconstruction quality varies when using input viewpoints (Sec.~\ref{sec:metrics} second column from the right).
We quantify this variation as the average standard deviation of performance on the same evaluation view using different input views, \ie vertical columns $1\leq j\leq N$ of the score matrix $S$ (Tab.~\ref{table:ablation} first column from the right).  
\begin{align}
IVV = Mean(\{Stddev(\{\textbf{S}_{t,i,j} | i\neq j\})\}  \notag \\ 
    | 1\leq j\leq N, 1\leq t\leq T\}) .
\end{align}
We observe that the inclusion of the Undistorter module consistently improves the "Novel View Variation" and "Input Robustness" metrics versus LP3D, indicating better robustness to different input viewpoints and more consistent rendering quality across views. 
However, when only the Undistorter is added (Tab.~\ref{table:ablation} second row) the PSNR is reduced. 
This is likely because this model does not leverage the reference image to improve the reconstruction of occluded areas.
Additionally, by only undistorting the reconstruction, the Undistorter-only model loses the ability to achieve higher average score (but also higher standard deviations) by simply overfitting to the input view.
Similarly, the Fuser-only (Tab.~\ref{table:ablation} second row) achieves better robustness to different input viewpoints and more consistent rendering quality across views, but lower PSNR score.
A likely cause is that without the Undistorter, the Fuser needs to overcome the challenge of fusing highly misaligned triplanes, where the person look drastically different in the raw triplane $T_{raw}$ and canonical triplane $T_{cano}$, possibly inducing more blurriness and alignment artifacts that lower the PSNR performane. 
Overall the best performance is achieved by including both the Undistorter and Fuser because the two modules complement each other. The Undistorter corrects the distortion in the raw triplane and thus reduces the challenges in fusing misaligned triplanes, and the Fuser recovers the occluded areas in the raw triplane.

\noindent\textbf{Qualitative results.} Due to limited space, we refer the readers to our supplementary for image and video results on real-world data. Extensive experiments show that our method achieves better temporal consistency than LP3D~\cite{trevithick2023} and more accurately captures dynamic information like expressions and lighting change than GPAvatar~\cite{chu2024gpavatar}.

\begin{table}[t]
\centering
\fontsize{8}{13}\selectfont 
\setlength\tabcolsep{4pt} 
\begin{tabularx}{\linewidth}{p{2.cm}cccc}
\toprule
 Method & PSNR$\uparrow$ & Input View (PSNR) $\downarrow$ & ID$\downarrow$ & Expr$\downarrow$  \\
\hline
 LP3D (orig.) & 18.721 & \underline{2.130} & 0.247 &  0.451\\
 LP3D (ours) & 22.331 & 1.025 & 0.168 & 0.215 \\
 w optical flow & 22.085 & 1.175 & 0.178 & 0.335\\
 w/o shoulder aug. & \underline{22.342} & \underline{0.829} & \textbf{0.153} & \underline{0.244} \\ \hline
  Ours & \textbf{22.770} & \textbf{0.245} & \underline{0.158} & \textbf{0.187} \\ \bottomrule 
\end{tabularx}
\vspace{-5pt}
\caption{\textbf{Other Ablations}: We ablate our LP3D to the original (row 1\&2), Undistorter compared to optical flow (row 3), and the effectiveness of shoulder augmentation (row 4).}
\label{tab:rebuttal_tab}
\end{table}

%% file: sec/5_conclusion.tex
\section{Discussion}
\label{sec:discussion}

\paragraph{Conclusion.}
Recognizing the individual limitations of per-frame single-view reconstruction and 3D reenactment methods, we presented the first single-view 3D lifting method to reconstruct a 3D photorealistic avatar with faithful dynamic information as well as temporal consistency, which marries the best of both worlds. We believe our method paves the way forward for creating a high-quality telepresence system accessible to consumers. 

\vspace{-3mm}

\paragraph{Limitations and future work.}
With our method, fusing an extreme side view with a very different expression to the reference view may result in blurry reconstruction due to ambiguity in triplane alignment. 
We use a single reference image, but incorporating multiple ones with different expressions and head poses could lead to further improvements.
While we focus on a modifying triplanes, tuning the feedforward network itself to integrate information across multiple temporal frames could lead to further improvements. 
Finally, due to the additional components, our current run-time performance is slower than real-time, which could be improved in future work.

%% file: main.bbl
\begin{thebibliography}{67}
\providecommand{\natexlab}[1]{#1}
\providecommand{\url}[1]{\texttt{#1}}
\expandafter\ifx\csname urlstyle\endcsname\relax
  \providecommand{\doi}[1]{doi: #1}\else
  \providecommand{\doi}{doi: \begingroup \urlstyle{rm}\Url}\fi

\bibitem[Athar et~al.(2022)Athar, Xu, Sunkavalli, Shechtman, and Shu]{athar2022rignerf}
ShahRukh Athar, Zexiang Xu, Kalyan Sunkavalli, Eli Shechtman, and Zhixin Shu.
\newblock Rignerf: Fully controllable neural 3d portraits.
\newblock In \emph{IEEE Conference on Computer Vision and Pattern Recognition (CVPR)}, 2022.

\bibitem[Bhattarai et~al.(2024)Bhattarai, Nie{\ss}ner, and Sevastopolsky]{bhattarai2024triplanenet}
Ananta~R. Bhattarai, Matthias Nie{\ss}ner, and Artem Sevastopolsky.
\newblock Triplanenet: An encoder for eg3d inversion.
\newblock 2024.

\bibitem[Buehler et~al.(2023)Buehler, Sarkar, Shah, Li, Wang, Helminger, Orts-Escolano, Lagun, Hilliges, Beeler, and Meka]{buehler2023preface}
Marcel~C. Buehler, Kripasindhu Sarkar, Tanmay Shah, Gengyan Li, Daoye Wang, Leonhard Helminger, Sergio Orts-Escolano, Dmitry Lagun, Otmar Hilliges, Thabo Beeler, and Abhimitra Meka.
\newblock Preface: A data-driven volumetric prior for few-shot ultra high-resolution face synthesis.
\newblock In \emph{IEEE International Conference on Computer Vision (ICCV)}, 2023.

\bibitem[Cao et~al.(2022)Cao, Simon, Kim, Schwartz, Zollhoefer, Saito, Lombardi, Wei, Belko, Yu, Sheikh, and Saragih]{cao2022authentic}
Chen Cao, Tomas Simon, Jin~Kyu Kim, Gabe Schwartz, Michael Zollhoefer, Shun-Suke Saito, Stephen Lombardi, Shih-En Wei, Danielle Belko, Shoou-I Yu, Yaser Sheikh, and Jason Saragih.
\newblock Authentic volumetric avatars from a phone scan.
\newblock \emph{ACM Transactions on Graphics (SIGGRAPH)}, 2022.

\bibitem[Chan et~al.(2022)Chan, Lin, Chan, Nagano, Pan, Mello, Gallo, Guibas, Tremblay, Khamis, Karras, and Wetzstein]{eg3d2022}
Eric~R. Chan, Connor~Z. Lin, Matthew~A. Chan, Koki Nagano, Boxiao Pan, Shalini~De Mello, Orazio Gallo, Leonidas Guibas, Jonathan Tremblay, Sameh Khamis, Tero Karras, and Gordon Wetzstein.
\newblock Efficient geometry-aware {3D} generative adversarial networks.
\newblock In \emph{IEEE Conference on Computer Vision and Pattern Recognition (CVPR)}, 2022.

\bibitem[Chu et~al.(2024)Chu, Li, Zeng, Yang, Lin, Liu, and Harada]{chu2024gpavatar}
Xuangeng Chu, Yu Li, Ailing Zeng, Tianyu Yang, Lijian Lin, Yunfei Liu, and Tatsuya Harada.
\newblock Gpavatar: Generalizable and precise head avatar from image(s).
\newblock In \emph{International Conference on Learning Representations (ICLR)}, 2024.

\bibitem[Deng et~al.(2019)Deng, Guo, Niannan, and Zafeiriou]{deng2018arcface}
Jiankang Deng, Jia Guo, Xue Niannan, and Stefanos Zafeiriou.
\newblock Arcface: Additive angular margin loss for deep face recognition.
\newblock In \emph{IEEE Conference on Computer Vision and Pattern Recognition (CVPR)}, 2019.

\bibitem[Deng et~al.(2022)Deng, Yang, Xiang, and Tong]{deng2022gram}
Yu Deng, Jiaolong Yang, Jianfeng Xiang, and Xin Tong.
\newblock Gram: Generative radiance manifolds for 3d-aware image generation.
\newblock In \emph{IEEE Conference on Computer Vision and Pattern Recognition (CVPR)}, 2022.

\bibitem[Doukas et~al.(2021)Doukas, Zafeiriou, and Sharmanska]{doukas2020headgan}
Michail~Christos Doukas, Stefanos Zafeiriou, and Viktoriia Sharmanska.
\newblock Headgan: One-shot neural head synthesis and editing.
\newblock In \emph{IEEE International Conference on Computer Vision (ICCV)}, 2021.

\bibitem[Drobyshev et~al.(2022)Drobyshev, Chelishev, Khakhulin, Ivakhnenko, Lempitsky, and Zakharov]{Drobyshev22MP}
Nikita Drobyshev, Jenya Chelishev, Taras Khakhulin, Aleksei Ivakhnenko, Victor Lempitsky, and Egor Zakharov.
\newblock Megaportraits: One-shot megapixel neural head avatars.
\newblock \emph{arXiv preprint arXiv:2207.07621}, 2022.

\bibitem[Feng et~al.(2021)Feng, Feng, Black, and Bolkart]{feng2021learning}
Yao Feng, Haiwen Feng, Michael~J Black, and Timo Bolkart.
\newblock Learning an animatable detailed 3d face model from in-the-wild images.
\newblock \emph{ACM Transactions on Graphics (ToG)}, 40\penalty0 (4):\penalty0 1--13, 2021.

\bibitem[Fr{\"u}hst{\"u}ck et~al.(2023)Fr{\"u}hst{\"u}ck, Sarafianos, Xu, Wonka, and Tung]{Fruehstueck2023VIVE3D}
Anna Fr{\"u}hst{\"u}ck, Nikolaos Sarafianos, Yuanlu Xu, Peter Wonka, and Tony Tung.
\newblock {VIVE3D}: Viewpoint-independent video editing using {3D-Aware GANs}.
\newblock In \emph{IEEE Conference on Computer Vision and Pattern Recognition (CVPR)}, 2023.

\bibitem[Gafni et~al.(2021)Gafni, Thies, Zollh{\"o}fer, and Nie{\ss}ner]{Gafni_2021_CVPR}
Guy Gafni, Justus Thies, Michael Zollh{\"o}fer, and Matthias Nie{\ss}ner.
\newblock Dynamic neural radiance fields for monocular 4d facial avatar reconstruction.
\newblock In \emph{IEEE Conference on Computer Vision and Pattern Recognition (CVPR)}, 2021.

\bibitem[Goodfellow et~al.(2014)Goodfellow, Pouget-Abadie, Mirza, Xu, Warde-Farley, Ozair, Courville, and Bengio]{goodfellow2014generative}
Ian Goodfellow, Jean Pouget-Abadie, Mehdi Mirza, Bing Xu, David Warde-Farley, Sherjil Ozair, Aaron Courville, and Yoshua Bengio.
\newblock Generative adversarial nets.
\newblock In \emph{Advances in Neural Information Processing Systems (NeurIPS)}, 2014.

\bibitem[Gu et~al.(2021)Gu, Liu, Wang, and Theobalt]{gu2021stylenerf}
Jiatao Gu, Lingjie Liu, Peng Wang, and Christian Theobalt.
\newblock {StyleNeRF}: {A} style-based {3D}-aware generator for high-resolution image synthesis.
\newblock \emph{arXiv preprint arXiv:2110.08985}, 2021.

\bibitem[Hong et~al.(2022{\natexlab{a}})Hong, Zhang, Shen, and Xu]{hong2022depth}
Fa-Ting Hong, Longhao Zhang, Li Shen, and Dan Xu.
\newblock Depth-aware generative adversarial network for talking head video generation.
\newblock 2022{\natexlab{a}}.

\bibitem[Hong et~al.(2022{\natexlab{b}})Hong, Peng, Xiao, Liu, and Zhang]{hong2021headnerf}
Yang Hong, Bo Peng, Haiyao Xiao, Ligang Liu, and Juyong Zhang.
\newblock Headnerf: A real-time nerf-based parametric head model.
\newblock In \emph{IEEE Conference on Computer Vision and Pattern Recognition (CVPR)}, 2022{\natexlab{b}}.

\bibitem[Jones et~al.(2009)Jones, Lang, Fyffe, Yu, Busch, McDowall, Bolas, and Debevec]{jones2009}
Andrew Jones, Magnus Lang, Graham Fyffe, Xueming Yu, Jay Busch, Ian McDowall, Mark Bolas, and Paul Debevec.
\newblock Achieving eye contact in a one-to-many 3d video teleconferencing system.
\newblock \emph{ACM Transactions on Graphics (SIGGRAPH)}, 2009.

\bibitem[Karras et~al.(2019)Karras, Laine, and Aila]{karras2019style}
Tero Karras, Samuli Laine, and Timo Aila.
\newblock A style-based generator architecture for generative adversarial networks.
\newblock In \emph{IEEE Conference on Computer Vision and Pattern Recognition (CVPR)}, 2019.

\bibitem[Kauff and Schreer(2002)]{Kauff2002}
Peter Kauff and Oliver Schreer.
\newblock An immersive 3d video-conferencing system using shared virtual team user environments.
\newblock In \emph{Proceedings of the 4th International Conference on Collaborative Virtual Environments}, 2002.

\bibitem[Kerbl et~al.(2023)Kerbl, Kopanas, Leimk{\"u}hler, and Drettakis]{kerbl3Dgaussians}
Bernhard Kerbl, Georgios Kopanas, Thomas Leimk{\"u}hler, and George Drettakis.
\newblock 3d gaussian splatting for real-time radiance field rendering.
\newblock \emph{ACM Transactions on Graphics (SIGGRAPH)}, 2023.

\bibitem[Kirschstein et~al.(2023)Kirschstein, Qian, Giebenhain, Walter, and Nie\ss{}ner]{Nersemble}
Tobias Kirschstein, Shenhan Qian, Simon Giebenhain, Tim Walter, and Matthias Nie\ss{}ner.
\newblock Nersemble: Multi-view radiance field reconstruction of human heads.
\newblock \emph{ACM Trans. Graph.}, 2023.

\bibitem[Ko et~al.(2023)Ko, Cho, Choi, Ryoo, and Kim]{ko20233d3dganinversion}
Jaehoon Ko, Kyusun Cho, Daewon Choi, Kwangrok Ryoo, and Seungryong Kim.
\newblock 3d gan inversion with pose optimization.
\newblock In \emph{IEEE Winter Conference on Applications of Computer Vision (WACV)}, 2023.

\bibitem[Lawrence et~al.(2021)Lawrence, Goldman, Achar, Blascovich, Desloge, Fortes, Gomez, Häberling, Hoppe, Huibers, Knaus, Kuschak, Martin-Brualla, Nover, Russell, Seitz, and Tong]{starline}
Jason Lawrence, Dan~B Goldman, Supreeth Achar, Gregory~Major Blascovich, Joseph~G. Desloge, Tommy Fortes, Eric~M. Gomez, Sascha Häberling, Hugues Hoppe, Andy Huibers, Claude Knaus, Brian Kuschak, Ricardo Martin-Brualla, Harris Nover, Andrew~Ian Russell, Steven~M. Seitz, and Kevin Tong.
\newblock Project starline: A high-fidelity telepresence system.
\newblock \emph{ACM Transactions on Graphics (SIGGRAPH ASIA)}, 2021.

\bibitem[Li et~al.(2017)Li, Bolkart, Black, Li, and Romero]{FLAME:SiggraphAsia2017}
Tianye Li, Timo Bolkart, Michael.~J. Black, Hao Li, and Javier Romero.
\newblock Learning a model of facial shape and expression from {4D} scans.
\newblock 2017.

\bibitem[Li et~al.(2023{\natexlab{a}})Li, Zhang, Wang, Zhao, Wang, Chen, Zhang, Wang, Bo, and Li]{li2023hidenerf}
Weichuang Li, Longhao Zhang, Dong Wang, Bin Zhao, Zhigang Wang, Mulin Chen, Bang Zhang, Zhongjian Wang, Liefeng Bo, and Xuelong Li.
\newblock One-shot high-fidelity talking-head synthesis with deformable neural radiance field.
\newblock In \emph{IEEE Conference on Computer Vision and Pattern Recognition (CVPR)}, 2023{\natexlab{a}}.

\bibitem[Li et~al.(2023{\natexlab{b}})Li, De~Mello, Liu, Nagano, Iqbal, and Kautz]{Li2023Oneshot}
Xueting Li, Shalini De~Mello, Sifei Liu, Koki Nagano, Umar Iqbal, and Jan Kautz.
\newblock Generalizable one-shot neural head avatar.
\newblock 2023{\natexlab{b}}.

\bibitem[Liang et~al.(2022)Liang, Fan, Xiang, Ranjan, Ilg, Green, Cao, Zhang, Timofte, and Van~Gool]{rvrt}
Jingyun Liang, Yuchen Fan, Xiaoyu Xiang, Rakesh Ranjan, Eddy Ilg, Simon Green, Jiezhang Cao, Kai Zhang, Radu Timofte, and Luc Van~Gool.
\newblock Recurrent video restoration transformer with guided deformable attention.
\newblock 2022.

\bibitem[Lin et~al.(2022)Lin, Lindell, Chan, and Wetzstein]{lin20223dganinversion}
C.Z. Lin, D.B. Lindell, E.R. Chan, and G. Wetzstein.
\newblock 3d gan inversion for controllable portrait image animation.
\newblock In \emph{ECCV Workshop on Learning to Generate 3D Shapes and Scenes}, 2022.

\bibitem[Ma et~al.(2021)Ma, Simon, Saragih, Wang, Li, Torre, and Sheikh]{PixelCodecAvatar}
S. Ma, T. Simon, J. Saragih, D. Wang, Y. Li, F.~La Torre, and Y. Sheikh.
\newblock Pixel codec avatars.
\newblock In \emph{IEEE Conference on Computer Vision and Pattern Recognition (CVPR)}, 2021.

\bibitem[Ma et~al.(2023)Ma, Zhu, Qi, Lei, and Zhang]{ma2023otavatar}
Zhiyuan Ma, Xiangyu Zhu, Guojun Qi, Zhen Lei, and Lei Zhang.
\newblock Otavatar: One-shot talking face avatar with controllable tri-plane rendering.
\newblock 2023.

\bibitem[Maimone et~al.(2012)Maimone, Bidwell, Peng, and Fuchs]{MAIMONE2012791}
Andrew Maimone, Jonathan Bidwell, Kun Peng, and Henry Fuchs.
\newblock Enhanced personal autostereoscopic telepresence system using commodity depth cameras.
\newblock \emph{Computers \& Graphics}, 2012.

\bibitem[Mildenhall et~al.(2020)Mildenhall, Srinivasan, Tancik, Barron, Ramamoorthi, and Ng]{mildenhall2020nerf}
Ben Mildenhall, Pratul~P Srinivasan, Matthew Tancik, Jonathan~T Barron, Ravi Ramamoorthi, and Ren Ng.
\newblock {NeRF}: {R}epresenting scenes as neural radiance fields for view synthesis.
\newblock In \emph{European Conference on Computer Vision (ECCV)}, 2020.

\bibitem[Or-El et~al.(2022)Or-El, Luo, Shan, Shechtman, Park, and Kemelmacher-Shlizerman]{orel2022stylesdf}
Roy Or-El, Xuan Luo, Mengyi Shan, Eli Shechtman, Jeong~Joon Park, and Ira Kemelmacher-Shlizerman.
\newblock Style{SDF}: {H}igh-{R}esolution {3D}-{C}onsistent {I}mage and {G}eometry {G}eneration.
\newblock In \emph{IEEE Conference on Computer Vision and Pattern Recognition (CVPR)}, 2022.

\bibitem[Orts-Escolano et~al.(2016)Orts-Escolano, Rhemann, Fanello, Chang, Kowdle, Degtyarev, Kim, Davidson, Khamis, Dou, Tankovich, Loop, Cai, Chou, Mennicken, Valentin, Pradeep, Wang, Kang, Kohli, Lutchyn, Keskin, and Izadi]{orts-escolano2016holoportation}
Sergio Orts-Escolano, Christoph Rhemann, Sean Fanello, Wayne Chang, Adarsh Kowdle, Yury Degtyarev, David Kim, Philip~L. Davidson, Sameh Khamis, Mingsong Dou, Vladimir Tankovich, Charles Loop, Qin Cai, Philip~A. Chou, Sarah Mennicken, Julien Valentin, Vivek Pradeep, Shenlong Wang, Sing~Bing Kang, Pushmeet Kohli, Yuliya Lutchyn, Cem Keskin, and Shahram Izadi.
\newblock Holoportation: Virtual 3d teleportation in real-time.
\newblock In \emph{UIST 2016}, 2016.

\bibitem[Park et~al.(2021)Park, Sinha, Barron, Bouaziz, Goldman, Seitz, and Martin-Brualla]{park2021nerfies}
Keunhong Park, Utkarsh Sinha, Jonathan~T. Barron, Sofien Bouaziz, Dan~B Goldman, Steven~M. Seitz, and Ricardo Martin-Brualla.
\newblock Nerfies: Deformable neural radiance fields.
\newblock 2021.

\bibitem[Pumarola et~al.(2020)Pumarola, Corona, Pons-Moll, and Moreno-Noguer]{pumarola2020d}
Albert Pumarola, Enric Corona, Gerard Pons-Moll, and Francesc Moreno-Noguer.
\newblock {D-NeRF: Neural Radiance Fields for Dynamic Scenes}.
\newblock In \emph{IEEE Conference on Computer Vision and Pattern Recognition (CVPR)}, 2020.

\bibitem[Qi et~al.(2023)Qi, Wu, Wang, Wang, and Sengupta]{qi2023my3dgen}
Luchao Qi, Jiaye Wu, Annie~N. Wang, Shengze Wang, and Roni Sengupta.
\newblock My3dgen: A scalable personalized 3d generative model, 2023.

\bibitem[Qian et~al.(2024)Qian, Kirschstein, Schoneveld, Davoli, Giebenhain, and Nie\ss{}ner]{qian2023gaussianavatars}
Shenhan Qian, Tobias Kirschstein, Liam Schoneveld, Davide Davoli, Simon Giebenhain, and Matthias Nie\ss{}ner.
\newblock Gaussianavatars: Photorealistic head avatars with rigged 3d gaussians.
\newblock 2024.

\bibitem[Ranjan and Black(2017)]{ranjan2017optical}
Anurag Ranjan and Michael~J Black.
\newblock Optical flow estimation using a spatial pyramid network.
\newblock In \emph{Proceedings of the IEEE conference on computer vision and pattern recognition}, pages 4161--4170, 2017.

\bibitem[Raskar et~al.(1998)Raskar, Welch, Cutts, Lake, Stesin, and Fuchs]{raskar1998office}
Ramesh Raskar, Greg Welch, Matt Cutts, Adam Lake, Lev Stesin, and Henry Fuchs.
\newblock The office of the future: a unified approach to image-based modeling and spatially immersive displays.
\newblock In \emph{ACM Transactions on Graphics (SIGGRAPH)}, 1998.

\bibitem[Rebain et~al.(2022)Rebain, Matthews, Yi, Lagun, and Tagliasacchi]{rebain2022lolnerf}
Daniel Rebain, Mark Matthews, Kwang~Moo Yi, Dmitry Lagun, and Andrea Tagliasacchi.
\newblock Lolnerf: Learn from one look.
\newblock In \emph{IEEE Conference on Computer Vision and Pattern Recognition (CVPR)}, 2022.

\bibitem[Saito et~al.(2024)Saito, Schwartz, Simon, Li, and Nam]{saito2024rgca}
Shunsuke Saito, Gabriel Schwartz, Tomas Simon, Junxuan Li, and Giljoo Nam.
\newblock Relightable gaussian codec avatars.
\newblock In \emph{IEEE Conference on Computer Vision and Pattern Recognition (CVPR)}, 2024.

\bibitem[Siarohin et~al.(2019)Siarohin, Lathuilière, Tulyakov, Ricci, and Sebe]{Siarohin_2019_NeurIPS}
Aliaksandr Siarohin, Stéphane Lathuilière, Sergey Tulyakov, Elisa Ricci, and Nicu Sebe.
\newblock First order motion model for image animation.
\newblock In \emph{Advances in Neural Information Processing Systems (NeurIPS)}, 2019.

\bibitem[Skorokhodov et~al.(2022)Skorokhodov, Tulyakov, Wang, and Wonka]{epigraf}
Ivan Skorokhodov, Sergey Tulyakov, Yiqun Wang, and Peter Wonka.
\newblock Epigraf: Rethinking training of 3d gans.
\newblock In \emph{Advances in Neural Information Processing Systems (NeurIPS)}, 2022.

\bibitem[Stengel et~al.(2023)Stengel, Nagano, Liu, Chan, Trevithick, De~Mello, Kim, and Luebke]{stengel20233dvc}
Michael Stengel, Koki Nagano, Chao Liu, Matthew Chan, Alex Trevithick, Shalini De~Mello, Jonghyun Kim, and David Luebke.
\newblock Ai-mediated 3d video conferencing.
\newblock In \emph{ACM SIGGRAPH 2023 Emerging Technologies}, 2023.

\bibitem[Sun et~al.(2022)Sun, Wang, Shi, Wang, Wang, and Liu]{sun2022ide}
Jingxiang Sun, Xuan Wang, Yichun Shi, Lizhen Wang, Jue Wang, and Yebin Liu.
\newblock Ide-3d: Interactive disentangled editing for high-resolution 3d-aware portrait synthesis.
\newblock \emph{ACM Transactions on Graphics (SIGGRAPH ASIA)}, 2022.

\bibitem[Sun et~al.(2023)Sun, Wang, Wang, Li, Zhang, Zhang, and Liu]{sun2023next3d}
Jingxiang Sun, Xuan Wang, Lizhen Wang, Xiaoyu Li, Yong Zhang, Hongwen Zhang, and Yebin Liu.
\newblock Next3d: Generative neural texture rasterization for 3d-aware head avatars.
\newblock In \emph{IEEE Conference on Computer Vision and Pattern Recognition (CVPR)}, 2023.

\bibitem[Tran et~al.(2024)Tran, Zakharov, Ho, Tran, Hu, and Li]{tran2023voodoo}
Phong Tran, Egor Zakharov, Long-Nhat Ho, Anh~Tuan Tran, Liwen Hu, and Hao Li.
\newblock Voodoo 3d: Volumetric portrait disentanglement for one-shot 3d head reenactment.
\newblock 2024.

\bibitem[Trevithick et~al.(2023)Trevithick, Chan, Stengel, Chan, Liu, Yu, Khamis, Chandraker, Ramamoorthi, and Nagano]{trevithick2023}
Alex Trevithick, Matthew Chan, Michael Stengel, Eric~R. Chan, Chao Liu, Zhiding Yu, Sameh Khamis, Manmohan Chandraker, Ravi Ramamoorthi, and Koki Nagano.
\newblock Real-time radiance fields for single-image portrait view synthesis.
\newblock In \emph{ACM Transactions on Graphics (SIGGRAPH)}, 2023.

\bibitem[Wang et~al.(2021)Wang, Mallya, and Liu]{wang2021facevid2vid}
Ting-Chun Wang, Arun Mallya, and Ming-Yu Liu.
\newblock One-shot free-view neural talking-head synthesis for video conferencing.
\newblock In \emph{IEEE Conference on Computer Vision and Pattern Recognition (CVPR)}, 2021.

\bibitem[Wang et~al.(2022)Wang, Yang, Bremond, and Dantcheva]{wang2022latent}
Yaohui Wang, Di Yang, Francois Bremond, and Antitza Dantcheva.
\newblock Latent image animator: Learning to animate images via latent space navigation.
\newblock In \emph{International Conference on Learning Representations (ICLR)}, 2022.

\bibitem[Xiang et~al.(2022)Xiang, Yang, Deng, and Tong]{xiang2022gramhd}
Jianfeng Xiang, Jiaolong Yang, Yu Deng, and Xin Tong.
\newblock Gram-hd: 3d-consistent image generation at high resolution with generative radiance manifolds.
\newblock \emph{arXiv preprint arXiv:2206.07255}, 2022.

\bibitem[Xu et~al.(2022)Xu, Peng, Yang, Shen, and Zhou]{xu2021volumegan}
Yinghao Xu, Sida Peng, Ceyuan Yang, Yujun Shen, and Bolei Zhou.
\newblock 3d-aware image synthesis via learning structural and textural representations.
\newblock In \emph{IEEE Conference on Computer Vision and Pattern Recognition (CVPR)}, 2022.

\bibitem[Xu et~al.(2024)Xu, Chen, Li, Zhang, Wang, Zheng, and Liu]{xu2023gaussianheadavatar}
Yuelang Xu, Benwang Chen, Zhe Li, Hongwen Zhang, Lizhen Wang, Zerong Zheng, and Yebin Liu.
\newblock Gaussian head avatar: Ultra high-fidelity head avatar via dynamic gaussians.
\newblock In \emph{IEEE Conference on Computer Vision and Pattern Recognition (CVPR)}, 2024.

\bibitem[Xu et~al.(2023)Xu, Zhang, Liew, Zhang, Bai, Feng, and Shou]{xu2022pv3d}
Zhongcong Xu, Jianfeng Zhang, Junhao Liew, Wenqing Zhang, Song Bai, Jiashi Feng, and Mike~Zheng Shou.
\newblock Pv3d: A 3d generative model for portrait video generation.
\newblock In \emph{The Tenth International Conference on Learning Representations}, 2023.

\bibitem[Ye et~al.(2024)Ye, Zhong, Ren, Yang, Li, Huang, Jiang, He, Huang, Liu, Zhang, Yin, Ma, and Zhao]{ye2024real3dportrait}
Zhenhui Ye, Tianyun Zhong, Yi Ren, Jiaqi Yang, Weichuang Li, Jiangwei Huang, Ziyue Jiang, Jinzheng He, Rongjie Huang, Jinglin Liu, Chen Zhang, Xiang Yin, Zejun Ma, and Zhou Zhao.
\newblock Real3d-portrait: One-shot realistic 3d talking portrait synthesis.
\newblock In \emph{International Conference on Learning Representations (ICLR)}, 2024.

\bibitem[Yin et~al.(2022)Yin, Zhang, Cun, Cao, Fan, Wang, Bai, Wu, Wang, and Yang]{StyleHeat2022}
Fei Yin, Yong Zhang, Xiaodong Cun, Mingdeng Cao, Yanbo Fan, Xuan Wang, Qingyan Bai, Baoyuan Wu, Jue Wang, and Yujiu Yang.
\newblock Styleheat: One-shot high-resolution editable talking face generation via pre-trained stylegan.
\newblock 2022.

\bibitem[Yu et~al.(2023)Yu, Fan, Zhang, Wang, Yin, Bai, Cao, Shan, Wu, Sun, and Wu]{NOFA2023Yu}
Wangbo Yu, Yanbo Fan, Yong Zhang, Xuan Wang, Fei Yin, Yunpeng Bai, Yan-Pei Cao, Ying Shan, Yang Wu, Zhongqian Sun, and Baoyuan Wu.
\newblock Nofa: Nerf-based one-shot facial avatar reconstruction.
\newblock In \emph{ACM SIGGRAPH 2023 Conference Proceedings}, 2023.

\bibitem[Zakharov et~al.(2019)Zakharov, Shysheya, Burkov, and Lempitsky]{Zakharov_2019_ICCV}
Egor Zakharov, Aliaksandra Shysheya, Egor Burkov, and Victor Lempitsky.
\newblock Few-shot adversarial learning of realistic neural talking head models.
\newblock In \emph{IEEE International Conference on Computer Vision (ICCV)}, 2019.

\bibitem[Zakharov et~al.(2020)Zakharov, Ivakhnenko, Shysheya, and Lempitsky]{Zakharov20}
Egor Zakharov, Aleksei Ivakhnenko, Aliaksandra Shysheya, and Victor Lempitsky.
\newblock Fast bi-layer neural synthesis of one-shot realistic head avatars.
\newblock In \emph{European Conference on Computer Vision (ECCV)}, 2020.

\bibitem[Zhang et~al.(2023)Zhang, Qi, Zhang, Zhang, Wu, Chen, Chen, Wang, and Wen]{zhang2022metaportrait}
Bowen Zhang, Chenyang Qi, Pan Zhang, Bo Zhang, HsiangTao Wu, Dong Chen, Qifeng Chen, Yong Wang, and Fang Wen.
\newblock Metaportrait: Identity-preserving talking head generation with fast personalized adaptation, 2023.

\bibitem[Zhang et~al.(2022)Zhang, Zheng, Gao, Zhang, Pan, and Yang]{zhang2022mvcgan}
Xuanmeng Zhang, Zhedong Zheng, Daiheng Gao, Bang Zhang, Pan Pan, and Yi Yang.
\newblock Multi-view consistent generative adversarial networks for 3d-aware image synthesis.
\newblock In \emph{IEEE Conference on Computer Vision and Pattern Recognition (CVPR)}, 2022.

\bibitem[Zhao and Zhang(2022)]{tps2022}
Jian Zhao and Hui Zhang.
\newblock Thin-plate spline motion model for image animation.
\newblock In \emph{IEEE Conference on Computer Vision and Pattern Recognition (CVPR)}, pages 3657--3666, 2022.

\bibitem[Zheng et~al.(2022)Zheng, Abrevaya, Bühler, Chen, Black, and Hilliges]{zheng2022imavatar}
Yufeng Zheng, Victoria~Fernández Abrevaya, Marcel~C. Bühler, Xu Chen, Michael~J. Black, and Otmar Hilliges.
\newblock {I} {M} {Avatar}: Implicit morphable head avatars from videos.
\newblock In \emph{IEEE Conference on Computer Vision and Pattern Recognition (CVPR)}, 2022.

\bibitem[Zhou et~al.(2021)Zhou, Xie, Ni, and Tian]{Zhou2021CIPS3D}
Peng Zhou, Lingxi Xie, Bingbing Ni, and Qi Tian.
\newblock {{CIPS}}-{{3D}}: A {{3D}}-{{Aware Generator}} of {{GANs Based}} on {{Conditionally}}-{{Independent Pixel Synthesis}}.
\newblock \emph{arXiv preprint arXiv:2110.09788}, 2021.

\bibitem[Zhuang et~al.(2022)Zhuang, Zhu, Sun, and Cao]{zhuang2022mofanerf}
Yiyu Zhuang, Hao Zhu, Xusen Sun, and Xun Cao.
\newblock Mofanerf: Morphable facial neural radiance field.
\newblock In \emph{European Conference on Computer Vision (ECCV)}, 2022.

\end{thebibliography}
